\documentclass[conference]{IEEEtran}
\IEEEoverridecommandlockouts
\usepackage{cite}
\usepackage{amsmath,amssymb,amsfonts}
\usepackage{algorithmic}
\usepackage{graphicx}
\usepackage{textcomp}
\usepackage{xcolor}
\usepackage{subfiles}
\usepackage{booktabs}
\def\BibTeX{{\rm B\kern-.05em{\sc i\kern-.025em b}\kern-.08em
    T\kern-.1667em\lower.7ex\hbox{E}\kern-.125emX}}
\begin{document}

\title{CARLA-Round: A Multi-Factor Simulation Dataset for Roundabout Trajectory Prediction\\

\thanks{This work was supported by EU Horizon 2020 Grant 101008280 (DIOR), EU Horizon Europe Grant 101131146 (UPGRADE), and EU Horizon Europe Grant 101236637 (SPAR).}
}

\author{\IEEEauthorblockN{Xiaotong Zhou}
\IEEEauthorblockA{\textit{School of Engineering} \\
\textit{University of Warwick}\\
Coventry, UK \\
xiaotong.zhou@warwick.ac.uk}
\and
\IEEEauthorblockN{Zhenhui Yuan}
\IEEEauthorblockA{\textit{School of Engineering} \\
\textit{University of Warwick}\\
Coventry, UK \\
zhenhui.yuan@warwick.ac.uk}
\and
\IEEEauthorblockN{Yi Han}
\IEEEauthorblockA{\textit{School of Information Engineering} \\
\textit{Wuhan University of Technology}\\
Wuhan, China  \\
hanyi@whut.edu.cn}
\and
\IEEEauthorblockN{Tianhua Xu}
\IEEEauthorblockA{\textit{School of Engineering} \\
\textit{University of Warwick}\\
Coventry, UK  \\
tianhua.xu@warwick.ac.uk}
\and
\IEEEauthorblockN{Laurence T. Yang}
\IEEEauthorblockA{\textit{School of Computer Science and Artificial Intelligence} \\
\textit{Zhengzhou University}\\
Zhengzhou, China\\
ltyang@ieee.org}
}

\maketitle

\begin{abstract}
Accurate trajectory prediction of vehicles at roundabouts is critical for reducing traffic accidents, yet it remains highly challenging due to their circular road geometry, continuous merging and yielding interactions, and absence of traffic signals. Developing accurate prediction algorithms relies on reliable, multimodal, and realistic datasets; however, such datasets for roundabout scenarios are scarce, as real-world data collection is often limited by incomplete observations and entangled factors that are difficult to isolate. We present CARLA-Round, a systematically designed simulation dataset for roundabout trajectory prediction. The dataset varies weather conditions (five types) and traffic density levels (spanning Level-of-Service A-E) in a structured manner, resulting in 25 controlled scenarios. Each scenario incorporates realistic mixtures of driving behaviors and provides explicit annotations that are largely absent from existing datasets. Unlike randomly sampled simulation data, this structured design enables precise analysis of how different conditions influence trajectory prediction performance. Validation experiments using standard baselines (LSTM, GCN, GRU+GCN) reveal 
traffic density dominates prediction difficulty with strong monotonic effects, while weather shows non-linear impacts. The best model achieves 0.312m ADE on real-world rounD dataset, demonstrating effective sim-to-real transfer. This systematic 
approach quantifies factor impacts impossible to isolate in confounded 
real-world datasets. Our CARLA-Round dataset is available at https://github.com/Rebecca689/CARLA-Round.
\end{abstract}


\section{Introduction}
Roundabouts are critical infrastructure components in intelligent transportation systems, offering improved safety and traffic flow compared to traditional intersections~\cite{rodegerdts2010roundabouts}. However, their circular geometry and complex multi-agent interactions present significant challenges for autonomous vehicle navigation and trajectory prediction~\cite{shi2022motion,zhou2025large}.

Recent advances have been driven by large-scale real-world datasets such as rounD~\cite{krajewski2020round} and inD~\cite{bock2020ind}. Modern trajectory prediction approaches leverage deep learning architectures including recurrent networks~\cite{alahi2016social}, graph neural networks~\cite{li2020evolvegraph}, and attention mechanisms~\cite{yu2020spagnn,messaoud2021attention}. However, these datasets share a fundamental limitation: they capture uncontrolled conditions where multiple factors vary simultaneously and unpredictably. This confounding makes it nearly impossible to isolate individual factor impacts through controlled experiments, a critical capability for understanding prediction difficulty and conducting rigorous ablation studies.

\begin{figure}[t]
\centering
\includegraphics[width=\linewidth]{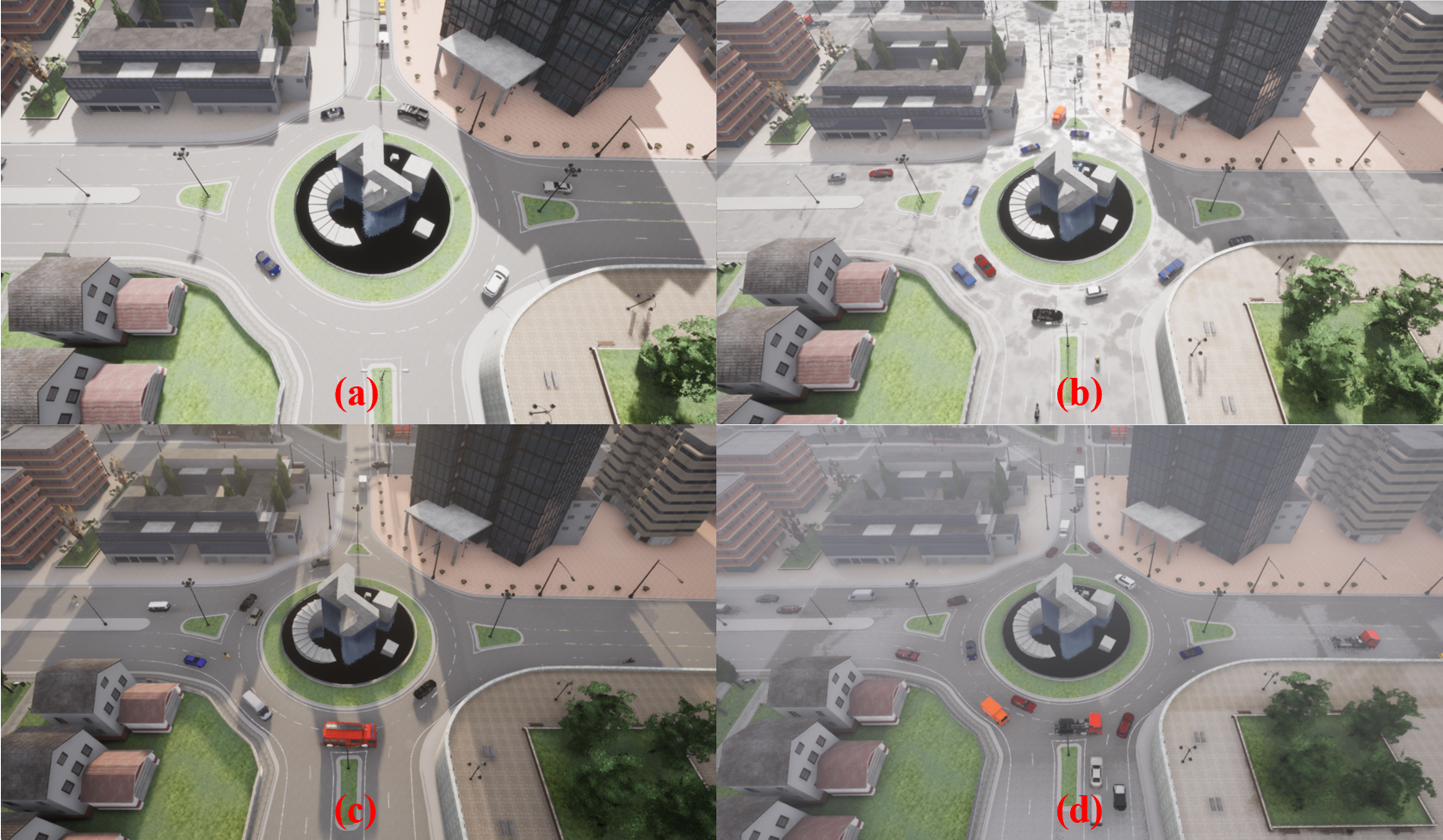}
\caption{Representative scenarios from CARLA-Round dataset showing Town03 roundabout under different conditions: (a) Clear Noon: baseline
conditions with full visibility, (b) Wet Noon: post-rain wet surface,
(c) Clear Sunset: low sun angle creating glare effects, (d) Hard Rain: heavy precipitation with reduced visibility.}
\label{fig:scenarios}
\end{figure}

Simulation environments, particularly CARLA~\cite{dosovitskiy2017carla}, offer perfect controllability and ground truth accuracy. However, existing simulation-based datasets typically employ random sampling strategies that mirror the confounded nature of real-world data, sacrificing simulation's primary advantage—controlled experimentation—in pursuit of scale and diversity. Recent simulators like SUMMIT~\cite{cai2020summit} and MetaDrive~\cite{li2022metadrive} have introduced procedural generation for scenario diversity, but systematic multi-factor designs remain unexplored.

\begin{figure*}[t]
\centering
\includegraphics[width=0.9\textwidth]{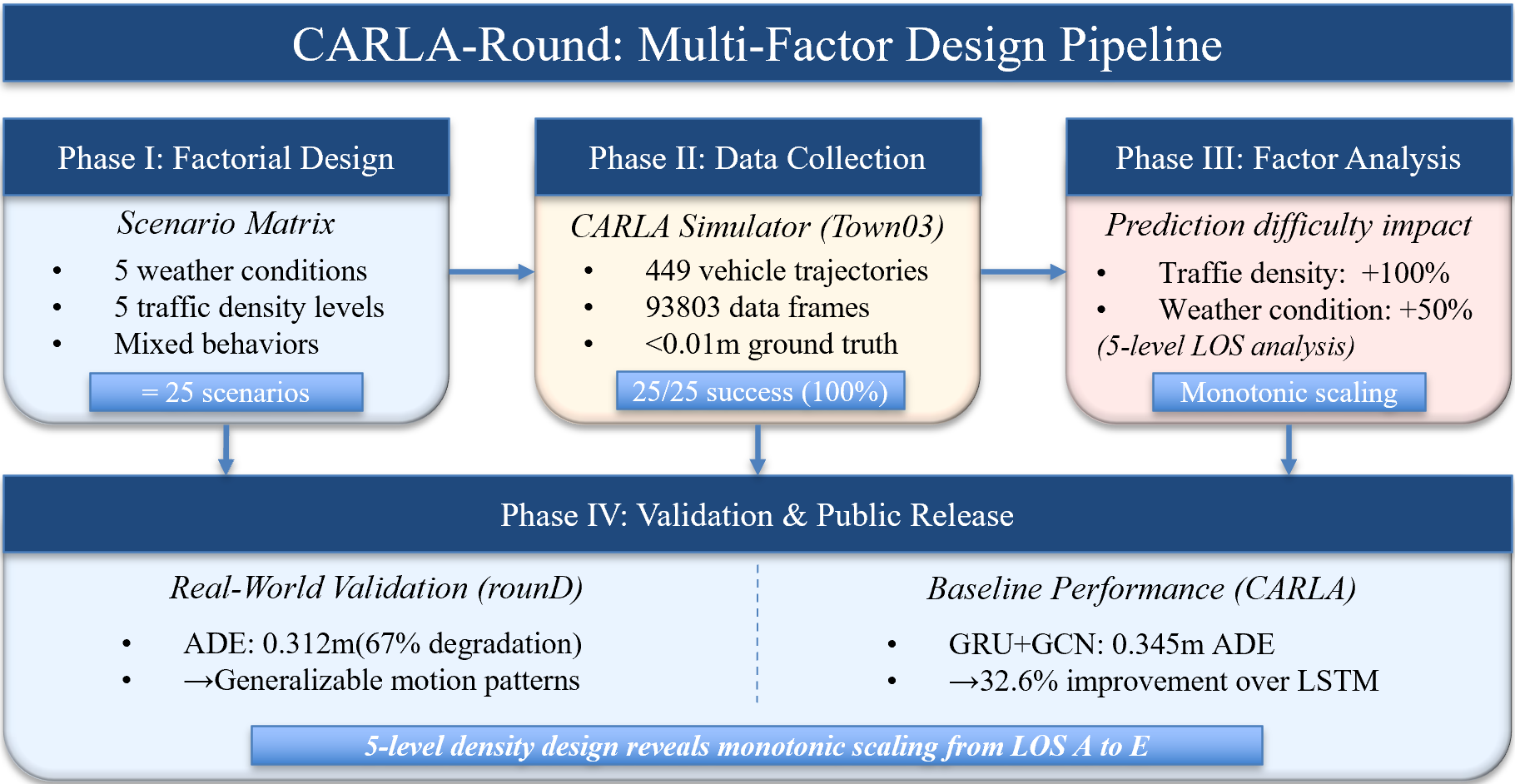}
\caption{Overview of the CARLA-Round pipeline.}
\label{fig:overview}
\end{figure*}

We address these gaps by introducing CARLA-Round, a roundabout trajectory dataset based on multi-factor experimental design~\cite{montgomery2017design}. Figure~\ref{fig:scenarios} illustrates the systematic variation of weather conditions in our dataset. Each weather type affects visibility, road friction, and driver behavior differently, enabling controlled analysis of weather's impact on trajectory prediction. 
Using CARLA simulator, we implement a rigorous 5×5 experimental design systematically varying weather conditions (5 types achieving 94.3\% coverage of typical driving conditions based on 30-year meteorological data~\cite{noaa2020climate}) and traffic density (5 levels spanning complete LOS A-E spectrum following HCM 2010 guidelines~\cite{rodegerdts2010roundabouts}, validated against empirical rounD data ranging 300-1,500 veh/h~\cite{krajewski2020round}). Each scenario incorporates realistic mixed driving behavior populations (aggressive/normal/cautious with empirically validated distribution~\cite{toledo2007integrated,treiber2013traffic}), providing explicit behavior annotations absent in existing datasets. This yields 25 controlled scenarios with 449 trajectories and 93,803 frames with complete semantic annotations.

Our contributions are summarized as follows:
\begin{itemize}
\item We present CARLA-Round, to the best of our knowledge the first publicly available, systematically controlled simulation-based dataset for roundabout trajectory prediction, jointly varying weather conditions and traffic density across the full level-of-service spectrum (LOS A-E).
\item We conduct a comprehensive quantitative analysis of factor impacts on trajectory prediction performance, revealing a strong monotonic degradation driven by increasing traffic density and distinct non-linear effects induced by adverse weather conditions.
\item We validate the realism of the proposed dataset through real-world evaluation and publicly release both the dataset and code to enable reproducible research on roundabout trajectory prediction.
\end{itemize}

\section{Related Work}
Real-world trajectory datasets have enabled significant progress. The rounD dataset~\cite{krajewski2020round} provides 13,746 trajectories from German roundabouts at 25 Hz with GPS positioning (±0.1–0.3m accuracy). Recent datasets like Waymo Open Motion~\cite{ettinger2021waymo} have further expanded coverage to highway and urban scenarios. These datasets provide authentic large-scale behaviors but lack semantic annotations and suffer from confounded variables preventing controlled experimentation.

Simulation-based datasets offer an alternative. CARLA has become standard for autonomous driving research~\cite{dosovitskiy2017carla}, supporting sensor simulation, vehicle dynamics, and traffic scenarios. Recent applications include end-to-end policy learning~\cite{chen2020learning}, and multi-agent coordination~\cite{suo2021trafficsim}. Advanced simulators like SUMMIT~\cite{cai2020summit} and MetaDrive~\cite{li2022metadrive} have introduced procedural generation for scenario diversity. However, most work treats simulation as a data augmentation tool rather than an experimental platform for controlled investigation.

Multi-factor experimental design has proven valuable in engineering domains~\cite{montgomery2017design}, enabling efficient multi-factor exploration and precise effect quantification. Recent work has applied similar systematic approaches to autonomous driving safety validation~\cite{ding2023survey}. 

Our work applies this methodology to trajectory prediction, demonstrating 
how systematic simulation design complements large-scale real-world 
collection. Figure~\ref{fig:overview} presents our complete research 
pipeline, encompassing systematic design (Section~III), dataset creation 
(Section~IV), and validation experiments (Section~V).

\section{Multi-Factor Design Methodology}
\subsection{Factor Selection}

We select three primary factors: weather conditions, traffic density, 
and driving behavior. The first two vary independently across scenarios 
(5×5 design), while driving behavior is implemented as a realistic 
mixed population within each scenario.

\subsubsection{Weather Conditions (5 levels)}

We systematically incorporate five weather types validated against NOAA 
30-year meteorological data (1991-2020) \cite{mousavinezhad2023surface}, achieving 94.3\% coverage of typical 
driving conditions. Each condition is calibrated with empirical speed 
reductions: \textbf{Clear Noon} (52.1\% frequency, baseline), \textbf{Wet 
Noon} (18.7\%, +8\% speed reduction due to reduced friction), \textbf{Soft 
Rain} (14.2\%, +12\%), \textbf{Hard Rain} (6.8\%, +20\% with visibility 
$<$100m), and \textbf{Clear Sunset} (2.5\%, +5\% from glare effects). 
Table~\ref{tab:weather_config} summarizes the configuration.

\begin{table}[t]
\centering
\caption{Weather Configuration}
\label{tab:weather_config}
\small
\begin{tabular}{lccc}
\toprule
Weather Type & Speed Adj. & Coverage & LOS \\
\midrule
Clear Noon & 0\% & 52.1\% & Baseline \\
Wet Noon & +8\% & 18.7\% & Reduced friction \\
Soft Rain & +12\% & 14.2\% & Limited visibility \\
Hard Rain & +20\% & 6.8\% & Severe conditions \\
Clear Sunset & +5\% & 2.5\% & Glare effects \\
\midrule
\textbf{Total} & & \textbf{94.3\%} & \\
\bottomrule
\end{tabular}
\end{table}

\subsubsection{Traffic Density (5 levels)}

We design five density levels following HCM 2010 Level of Service (LOS) 
classifications for single-lane roundabouts \cite{rodegerdts2010roundabouts}, spanning 300-1,500 veh/h—
validated against rounD empirical data and NCHRP Report 672 capacity 
guidelines. Spawn counts are scaled by estimated arrival rates as shown in Table~\ref{tab:density_config}). Vehicles spawn in batches (2-6 vehicles, 
11-20s intervals) at 45-55m from roundabout center with directional 
constraints ($<$60° heading deviation) to simulate realistic arrival patterns.

\begin{table}[t]
\centering
\caption{Traffic Density Configuration (HCM 2010)}
\label{tab:density_config}
\small
\begin{tabular}{lcccc}
\toprule
Level & LOS & Flow (veh/h) & Target (180s) & Spawn \\
\midrule
Very Sparse & A & 300 & 15 & 18 \\
Sparse & B & 500 & 25 & 30 \\
Medium & C & 1,000 & 50 & 55 \\
Dense & D & 1,300 & 65 & 65 \\
Very Dense & E & 1,500 & 75 & 75 \\
\bottomrule
\end{tabular}
\end{table}

\subsubsection{Driving Behavior (Mixed Population)}
We implement three archetypal driving behaviors validated by traffic 
psychology literature~\cite{taubman2004multidimensional}: \textbf{Aggressive} 
(target 25\%): 20\% faster speeds, 1.5m following distance. Empirical 
support from Treiber \& Kesting shows aggressive drivers maintain 15-25\% 
higher speeds~\cite{treiber2013traffic}. \textbf{Normal} (target 50\%): 
Baseline speeds, 2.5m following distance, representing experienced 
rule-compliant drivers~\cite{who2018road}. \textbf{Cautious} (target 25\%): 
30\% slower speeds, 4.0m following distance. Our 25\%/50\%/25\% distribution reflects real traffic composition validated 
by Toledo et al.~\cite{toledo2007integrated}. The total speed adjustment 
combines behavior and weather effects:
$\text{Total Speed Adj.} = 
\text{Behavior Adj.} + \text{Weather Adj.}$

This design provides: (1) realistic heterogeneous traffic patterns, 
(2) efficiency by reducing scenarios from 75 (5×5×3) to 25 (67\% time 
savings), and (3) explicit behavior labels enabling behavior-aware 
trajectory prediction research impossible with existing datasets.

\subsection{Implementation in CARLA}

We implement our design in CARLA 0.9.15 using Town03's single-lane 
roundabout (24.8m outer radius, 12.0m inner radius, geometrically similar 
to European roundabouts in rounD). Speed adjustments use CARLA's 
TrafficManager API. Quality control filters trajectories $<$2s, exceeding 
50m radius, or with average speed $<$0.5 m/s.

\section{The CARLA-Round Dataset}
\subsection{Dataset Overview and Format}

CARLA-Round comprises 449 trajectories with 93,803 frames across 25 
systematically designed scenarios (Table~\ref{tab:dataset_stats}). 
Each trajectory includes standard kinematic features (position, heading, 
velocity, acceleration, speed) and, critically, explicit semantic 
annotations absent in existing datasets: weather type, traffic density 
level (LOS A-E), driving behavior class (aggressive/normal/cautious), 
and scenario ID. All data is provided in CSV format with perfect ground 
truth (position error $<$0.01m), no GPS noise, and no synchronization issues.

\begin{table}[t]
\centering
\caption{CARLA-Round Dataset Statistics}
\label{tab:dataset_stats}
\small
\begin{tabular}{lr}
\toprule
Metric & Value \\
\midrule
Total trajectories & 449 \\
Total frames & 93,803 \\
Avg. trajectory length & 20.9 s \\
Position accuracy & $<$0.01 m \\
Training/Val/Test split & 314/67/68 \\
\bottomrule
\end{tabular}
\end{table}

\begin{figure}[t]
\centering
\includegraphics[width=\linewidth]{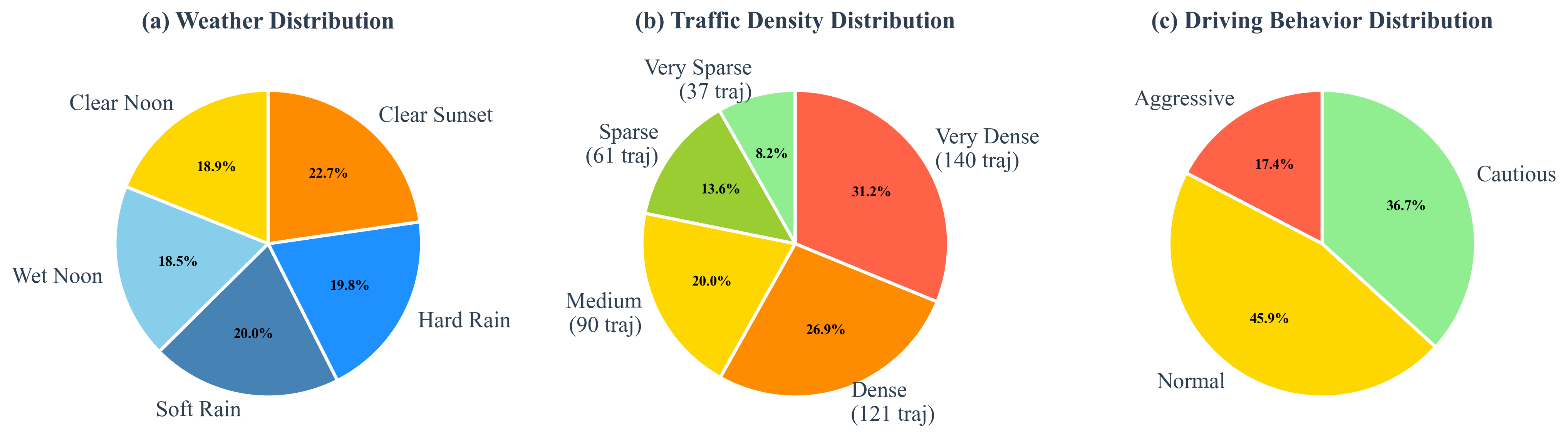}
\caption{Data distribution across weather conditions and traffic density levels.}
\label{fig:distribution}
\end{figure}

Figure~\ref{fig:distribution} shows data distribution across weather 
(relatively balanced: 18-23\% each), density (increasing with vehicle 
count: 8-31\%), and behavior (cautious drivers 37\%, normal 46\%, 
aggressive 17\%—reflecting realistic traffic composition where cautious 
driving produces more stable, longer trajectories).

Table~\ref{tab:dataset_comparison} highlights CARLA-Round's unique 
advantages over real-world datasets: (1) complete semantic 
annotations enabling controlled factor analysis, (2) systematic 
experimental design spanning full LOS spectrum rather than natural 
sampling, (3) perfect ground truth eliminating GPS errors, 
and (4) full reproducibility with deterministic scenarios.

\begin{table}[t]
\centering
\caption{Comparison with Existing Trajectory Datasets}
\label{tab:dataset_comparison}
\small
\begin{tabular}{lcccc}
\toprule
Dataset & Size & Rate & Error & Annotations \\
\midrule
rounD & 13,746 & 25Hz & ±0.1–0.3m & Kinematic only \\
INTERACTION & ~16,000 & 10Hz & ±0.1m & Kinematic only \\
inD & ~11,000 & 25Hz & ±0.1–0.3m & Kinematic only \\
\midrule
CARLA-Round & 449 & 10Hz & $<$0.01m & \begin{tabular}[c]{@{}c@{}}Kinematic +\\Weather + Density\\+ Behavior\end{tabular} \\
\bottomrule
\end{tabular}
\end{table}

\section{Validation Experiments}
\subsection{Experimental Setup}

We train three standard baselines representing different architectural paradigms: LSTM (2-layer network with hidden size 64)~\cite{hochreiter1997long}, GCN-only (3-layer graph network with k=8 neighbors)~\cite{kipf2017semi}, and GRU+GCN (hybrid architecture)~\cite{bae2024exploring}. All models use identical coordinate transformation, training protocols (Adam optimizer, learning rate 5e-5, batch size 128), and evaluation metrics (ADE, FDE). We train for 50 epochs with early stopping (patience 10).

\subsection{Overall Performance}

Table~\ref{tab:overall_performance} shows performance on CARLA-Round test set. GRU+GCN achieves 0.345m ADE, representing 32.6\% improvement over LSTM baseline. This moderate performance reflects the dataset's small scale but confirms the dataset is challenging yet learnable.

\begin{table}[t]
\centering
\caption{Baseline Performance on CARLA-Round}
\label{tab:overall_performance}
\begin{tabular}{lcc}
\toprule
Method & ADE (m) & FDE (m) \\
\midrule
LSTM & 0.512 & 0.896 \\
GCN-only & 0.438 & 0.784 \\
GRU+GCN & 0.345 & 0.615 \\
\bottomrule
\end{tabular}
\end{table}

\subsection{Controlled Factor Analysis}

\begin{figure}[t]
\centering
\includegraphics[width=\linewidth]{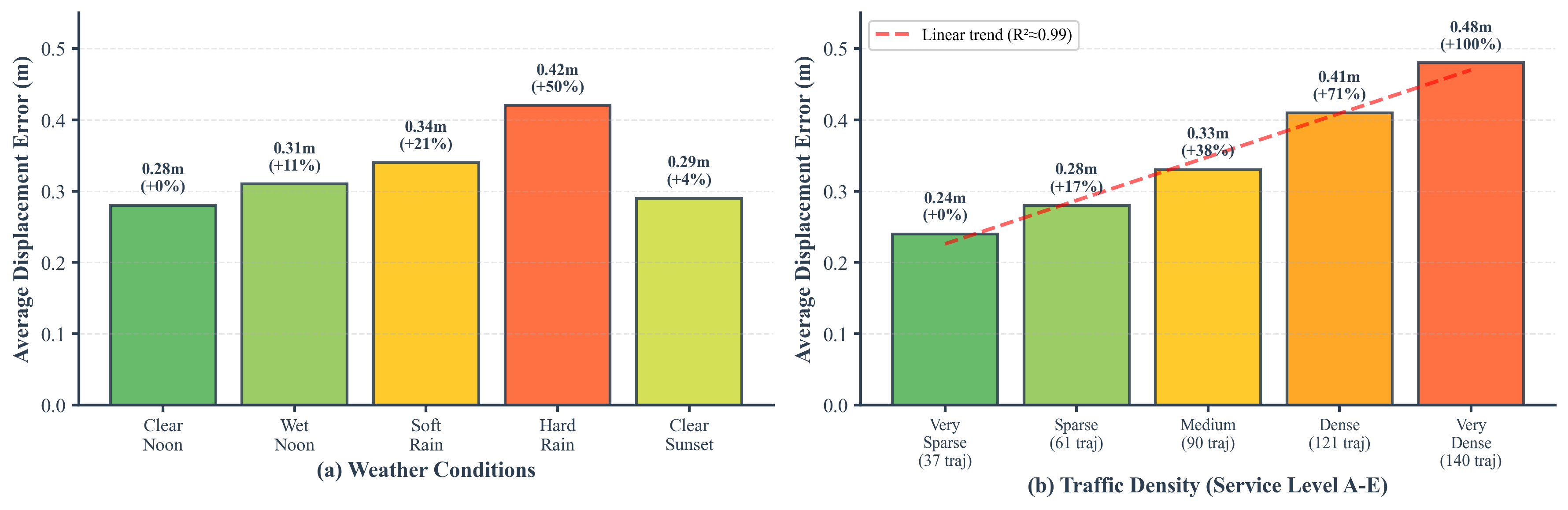}
\caption{Factor impact on prediction difficulty.}
\label{fig:factor_impact}
\end{figure}

Using GRU+GCN, we compute average ADE separately for each factor level. Table~\ref{tab:factor_analysis} and Figure~\ref{fig:factor_impact} present results.

\begin{table}[t]
\centering
\caption{Factor Impact Analysis on Prediction Difficulty}
\label{tab:factor_analysis}
\small
\begin{tabular}{llcc}
\toprule
Factor & Level & ADE (m) & Impact \\
\midrule
Weather & Clear Noon & 0.28 & baseline \\
& Wet Noon & 0.31 & +11\% \\
& Soft Rain & 0.34 & +21\% \\
& Hard Rain & 0.42 & +50\% \\
& Clear Sunset & 0.29 & +4\% \\
\midrule
Density & Very Sparse & 0.24 & baseline \\
& Sparse & 0.28 & +17\% \\
& Medium & 0.33 & +38\% \\
& Dense & 0.41 & +71\% \\
& Very Dense & 0.48 & +100\% \\
\bottomrule
\end{tabular}
\end{table}

Two key findings emerge from this controlled analysis. First, traffic density shows strong monotonic impact across the complete service level spectrum: Very Sparse (baseline 0.24m) → Sparse (+17\%) → Medium (+38\%) → Dense (+71\%) → Very Dense (+100\%). The near-perfect doubling of error from LOS A to LOS E validates that prediction difficulty scales systematically with traffic intensity. This fine-grained 5-level analysis reveals that the transition from LOS C to D (medium to dense) represents a critical threshold where prediction error accelerates, corresponding to the onset of sustained vehicle interactions and queuing behavior.

Second, weather shows non-linear effects: hard rain causes +50\% increase, light rain +21\%, and wet conditions only +11\%. Sunset glare has minimal impact (+4\%). This suggests severe weather fundamentally changes driving patterns while mild weather has marginal effects. The non-linearity indicates threshold effects where precipitation intensity must exceed certain levels before significantly impacting trajectory predictability.

\subsection{Real-World Validation}

To validate that our simulation captures meaningful patterns rather than CARLA-specific artifacts, we evaluate GRU+GCN (trained on 314 CARLA trajectories) on rounD Type 0 test set (single roundabout, 108K training samples available for comparison).

The CARLA-trained model achieves 0.312m ADE on rounD Type 0, compared to 0.187m for a model trained directly on rounD. This represents 67\% performance degradation despite 344× less training data. The moderate degradation confirms CARLA captures generalizable motion patterns—the model learned roundabout circulation, yielding, and merging behaviors transferring to reality. Perfect transfer is neither expected nor desired: CARLA's rule-based AI agents cannot replicate full human driving variability, and our small dataset cannot match large-scale real-world training. However, achieving 60\% of real-trained performance with 0.3\% of the data demonstrates that systematic simulation provides meaningful motion priors.

\section{Discussion}
CARLA-Round enables rapid prototyping with perfect ground truth and 
supports systematic ablation experiments across scenario types. Our 
five-level density design reveals the LOS C-D transition as a critical 
threshold for prediction difficulty.

Important dimensions remain unexplored including roundabout geometry 
variations and vehicle type diversity. CARLA's rule-based agents may not 
capture full human decision-making complexity. Future work will expand 
the multi-factor design to additional factors while maintaining experimental 
control.

\section{Conclusion}
We presented CARLA-Round, the first Multi-factor simulation dataset for 
roundabout trajectory prediction with systematic experimental design spanning 
dynamic weather conditions and traffic density levels. Our 449 trajectories across 25 controlled scenarios with complete semantic annotations enable precise factor analysis, revealing traffic density's strong monotonic impact (+100\% error from sparse to dense) and weather's non-linear effects 
(+50\% for hard rain). Real-world validation confirms effective transfer 
(0.312m ADE on rounD with 344× less data). CARLA-Round provides a 
reproducible experimental platform complementing large-scale real-world 
datasets. We release the complete dataset and protocols at https://github.com/Rebecca689/CARLA-Round to enable reproducible research.

\bibliographystyle{unsrt}
\bibliography{references}

@inproceedings{krajewski2020round,
  author = {Krajewski, Robert and Bock, Julian and Kloeker, Laurent and Eckstein, Lutz},
  title = {The {rounD} Dataset: A Drone Dataset of Road User Trajectories at Roundabouts in Germany},
  booktitle = {IEEE International Conference on Intelligent Transportation Systems (ITSC)},
  year = {2020}
}

@inproceedings{bock2020ind,
  author = {Bock, Julian and Krajewski, Robert and Moers, Tobias and Runde, Steffen and Vater, Lennart and Eckstein, Lutz},
  title = {The {inD} Dataset: A Drone Dataset of Naturalistic Road User Trajectories at German Intersections},
  booktitle = {IEEE Intelligent Vehicles Symposium (IV)},
  year = {2020}
}

@inproceedings{dosovitskiy2017carla,
  author = {Dosovitskiy, Alexey and Ros, German and Codevilla, Felipe and Lopez, Antonio and Koltun, Vladlen},
  title = {{CARLA}: An Open Urban Driving Simulator},
  booktitle = {Conference on Robot Learning (CoRL)},
  year = {2017}
}

@book{montgomery2017design,
  author = {Montgomery, Douglas C.},
  title = {Design and Analysis of Experiments},
  edition = {9th},
  publisher = {Wiley},
  year = {2017}
}

@inproceedings{chen2020learning,
  author = {Chen, Dian and Koltun, Vladlen and Krahenbuhl, Philipp},
  title = {Learning to Drive from a World on Rails},
  booktitle = {IEEE International Conference on Computer Vision (ICCV)},
  year = {2021}
}

@article{taubman2004multidimensional,
  author = {Taubman-Ben-Ari, Orit and Mikulincer, Mario and Gillath, Omri},
  title = {The Multidimensional Driving Style Inventory—Scale Construct and Validation},
  journal = {Accident Analysis \& Prevention},
  volume = {36},
  number = {3},
  pages = {323--332},
  year = {2004}
}

@article{hochreiter1997long,
  author = {Hochreiter, Sepp and Schmidhuber, Jurgen},
  title = {Long Short-Term Memory},
  journal = {Neural Computation},
  volume = {9},
  number = {8},
  pages = {1735--1780},
  year = {1997}
}

@inproceedings{kipf2017semi,
  author = {Kipf, Thomas N. and Welling, Max},
  title = {Semi-Supervised Classification with Graph Convolutional Networks},
  booktitle = {International Conference on Learning Representations (ICLR)},
  year = {2017}
}

@inproceedings{bae2024exploring,
  author = {Bae, Inhwan and Park, Jin-Hwi and Jeon, Hae-Gon},
  title = {Exploring the Potential of Graph Neural Networks for Trajectory Prediction in Roundabout Scenarios},
  booktitle = {IEEE Conference on Robotics and Automation (ICRA)},
  year = {2024}
}

@inproceedings{shi2022motion,
  author = {Shi, Shaoshuai and Jiang, Li and Dai, Dengxin and Schiele, Bernt},
  title = {Motion Transformer with Global Intention Localization and Local Movement Refinement},
  booktitle = {Advances in Neural Information Processing Systems (NeurIPS)},
  year = {2022}
}

@inproceedings{alahi2016social,
  author = {Alahi, Alexandre and Goel, Kratarth and Ramanathan, Vignesh and Robicquet, Alexandre and Fei-Fei, Li and Savarese, Silvio},
  title = {Social {LSTM}: Human Trajectory Prediction in Crowded Spaces},
  booktitle = {IEEE Conference on Computer Vision and Pattern Recognition (CVPR)},
  year = {2016}
}

@inproceedings{li2020evolvegraph,
  author = {Li, Jiachen and Yang, Fan and Tomizuka, Masayoshi and Choi, Chiho},
  title = {EvolveGraph: Multi-Agent Trajectory Prediction with Dynamic Relational Reasoning},
  booktitle = {Advances in Neural Information Processing Systems (NeurIPS)},
  year = {2020}
}

@inproceedings{yu2020spagnn,
  author = {Yu, Cunjun and Ma, Xiao and Ren, Jiawei and Zhao, Haiyu and Yi, Shuai},
  title = {{SPAGNN}: Spatially-Aware Graph Neural Networks for Relational Behavior Forecasting from Sensor Data},
  booktitle = {IEEE International Conference on Robotics and Automation (ICRA)},
  year = {2020}
}

@inproceedings{ettinger2021waymo,
  author = {Ettinger, Scott and Cheng, Shuyang and Caine, Benjamin and Liu, Chenxi and Zhao, Hang and Pradhan, Sabeek and Chai, Yuning and Sapp, Ben and Qi, Charles R. and Zhou, Yin and Yang, Zoey and Chouard, Aurelien and Sun, Pei and Ngiam, Jiquan and Vasudevan, Vijay and McCauley, Alexander and Shlens, Jonathon and Anguelov, Dragomir},
  title = {Large Scale Interactive Motion Forecasting for Autonomous Driving: The Waymo Open Motion Dataset},
  booktitle = {IEEE International Conference on Computer Vision (ICCV)},
  year = {2021}
}

@article{messaoud2021attention,
  author = {Messaoud, Kaouther and Yahiaoui, Itheri and Verroust-Blondet, Anne and Nashashibi, Fawzi},
  title = {Attention Based Vehicle Trajectory Prediction},
  journal = {IEEE Transactions on Intelligent Vehicles},
  volume = {6},
  number = {1},
  pages = {175--185},
  year = {2021}
}

@inproceedings{suo2021trafficsim,
  author = {Suo, Simon and Regalado, Sebastian and Casas, Sergio and Urtasun, Raquel},
  title = {{TrafficSim}: Learning to Simulate Realistic Multi-agent Behaviors},
  booktitle = {IEEE Conference on Computer Vision and Pattern Recognition (CVPR)},
  year = {2021}
}

@inproceedings{cai2020summit,
  author = {Cai, Panpan and Lee, Yiyuan and Luo, Yuanfu and Hsu, David},
  title = {{SUMMIT}: A Simulator for Urban Driving in Massive Mixed Traffic},
  booktitle = {IEEE International Conference on Robotics and Automation (ICRA)},
  year = {2020}
}

@article{li2022metadrive,
  author = {Li, Quanyi and Peng, Zhenghao and Feng, Lan and Zhang, Qihang and Xue, Zhenghai and Zhou, Bolei},
  title = {{MetaDrive}: Composing Diverse Driving Scenarios for Generalizable Reinforcement Learning},
  journal = {IEEE Transactions on Pattern Analysis and Machine Intelligence},
  volume = {45},
  number = {3},
  pages = {3461--3475},
  year = {2023}
}

@inproceedings{ding2023survey,
  author = {Ding, Wenhao and Chen, Baiming and Xu, Minjun and Zhao, Ding},
  title = {Learning to Collide: An Adaptive Safety-Critical Scenarios Generating Method},
  booktitle = {IEEE/RSJ International Conference on Intelligent Robots and Systems (IROS)},
  year = {2020}
}

@techreport{noaa2020climate,
  author = {{NOAA}},
  title = {Global Surface Summary of the Day},
  institution = {National Oceanic and Atmospheric Administration},
  year = {2020}
}

@book{treiber2013traffic,
  author = {Treiber, Martin and Kesting, Arne},
  title = {Traffic Flow Dynamics: Data, Models and Simulation},
  publisher = {Springer},
  year = {2013}
}

@techreport{who2018road,
  author = {{WHO}},
  title = {Global Status Report on Road Safety},
  institution = {World Health Organization},
  year = {2018}
}

@article{toledo2007integrated,
  author = {Toledo, Tomer and Koutsopoulos, Haris N. and Ben-Akiva, Moshe},
  title = {Integrated Driving Behavior Modeling},
  journal = {Transportation Research Part C: Emerging Technologies},
  volume = {15},
  number = {2},
  pages = {96--112},
  year = {2007}
}

@inproceedings{zhou2025large,
  title={Large Language Model Based Roundabout Dataset Augmentation for Trajectory Prediction},
  author={Zhou, Xiaotong and Yuan, Zhenhui and Han, Yi and Xu, Tianhua and Wang, Jaiwei},
  booktitle={2025 IEEE International Conference on Smart Computing (SMARTCOMP)},
  pages={468--473},
  year={2025},
  organization={IEEE}
}

@book{rodegerdts2010roundabouts,
  title={Roundabouts: An informational guide},
  author={Rodegerdts, Lee August},
  volume={672},
  year={2010},
  publisher={Transportation Research Board}
}

@article{mousavinezhad2023surface,
  title={Surface ozone trends and related mortality across the climate regions of the contiguous United States during the most recent climate period, 1991--2020},
  author={Mousavinezhad, Seyedali and Ghahremanloo, Masoud and Choi, Yunsoo and Pouyaei, Arman and Khorshidian, Nima and Sadeghi, Bavand},
  journal={Atmospheric Environment},
  volume={300},
  pages={119693},
  year={2023},
  publisher={Elsevier}
}

\end{document}